\documentclass[prd, aps, amsfonts, amssymb, amsmath, showkeys, footinbib, notitlepage,superscriptaddress,floatfix,longbibliography]{revtex4-1}

\usepackage{graphicx}%
\usepackage{dcolumn}%
\usepackage{bm}%
\usepackage{float}
\usepackage{xcolor}
\usepackage{mhchem}
\usepackage{gensymb}

\usepackage{newtxtext}
\usepackage[slantedGreek]{newtxmath}

\usepackage{comment}
\usepackage[pdftex, bookmarks, pdffitwindow=false, pdfstartview=FitH, pdfdisplaydoctitle, colorlinks, plainpages=false, pdftitle={},pdfauthor={}, pdfpagelabels, hypertexnames, citecolor={blue!50!black},linkcolor={blue!50!black}, urlcolor={blue!50!black}, pdflang={en}, hyperfootnotes=false, breaklinks]{hyperref}

\usepackage{siunitx}

\makeatletter
\renewcommand{\@seccntformat}[1]{}
\makeatother

\makeatletter
\def\@bibdataout@aps{
 \immediate\write\@bibdataout{
 @CONTROL{
   apsrev41Control, author="48",editor="1",pages="0",title="0",year="1"
 }}
 \if@filesw
  \immediate\write\@auxout{\string\citation{apsrev41Control}}
 \fi
}
\makeatother

\begin{document}

\preprint{APS/123-QED}

\title{Response Matching for generating materials and molecules}

\author{Bingqing Cheng}
\email{bingqingcheng@berkeley.edu}
\affiliation{Department of Chemistry, University of California, Berkeley, CA, USA}
\affiliation{The Institute of Science and Technology Austria, Am Campus 1, 3400 Klosterneuburg, Austria}

\date{\today}%

\begin{abstract}
Machine learning has recently emerged as a powerful tool for generating new molecular and material structures. The success of state-of-the-art models stems from their ability to incorporate physical symmetries, such as translation, rotation, and periodicity. Here, we present a novel generative method called Response Matching (RM), which leverages the fact that each stable material or molecule exists at the minimum of its potential energy surface. Consequently, any perturbation induces a response in energy and stress, driving the structure back to equilibrium. Matching to such response is closely related to score matching in diffusion models. 
By employing the combination of a machine learning interatomic potential and random structure search as the denoising model, RM exploits the locality of atomic interactions, and inherently respects permutation, translation, rotation, and periodic invariances. 
RM is the first model to handle both molecules and bulk materials under the same framework.
We demonstrate the efficiency and generalization of RM across three systems: a small organic molecular dataset, stable crystals from the Materials Project, and one-shot learning on a single diamond configuration.
\end{abstract}

\maketitle

\section{Introduction}

The exploration of new materials and molecules is crucial for technological advancements. Previous approaches rely on human intuition to propose and synthesize new molecules or high-throughput computational screening~\cite{jain2013commentary}. While these methods have led to some of the most important discoveries in history, they are often time-consuming and expensive, restricting the chemical space explored. This has encouraged the use of machine learning to generate new molecular and material structures by training on datasets of equilibrium atomistic structures~\cite{bilodeau2022generative,anstine2023generative}. 

A crucial aspect of recent generative models for molecules is the encoding of geometric symmetries, such as translations and rotations. For materials, it's also essential to account for periodic boundary conditions (PBC) in the structures~\cite{zeni2023mattergen,xie2021crystal}. Moreover, efficiently account for different element types is vital, since materials can contain over a hundred elements from the periodic table.

However, current generative models of atomistic structures do not incorporate a key inductive bias: the locality of atomic interactions. According to the locality assumption, the energy and forces acting on an atom depend solely on its neighboring atoms within a specific cutoff radius. This assumption is crucial for the success of most state-of-the-art machine learning interatomic potentials (MLIPs)~\cite{behler2007generalized,bartok2010gaussian,shapeev2016moment,drautz2019atomic,batzner20223,batatia2022mace,cheng2024cartesian}.
Another physical bias not currently exploited is that atoms cannot get too close to each other due to strong repulsive forces at short interatomic distances.

Another key insight is that every stable material or molecule resides at the minimum of the potential energy surface (PES) of the system. 
As such, when slight perturbations are introduced to the atomic positions of the stable structure, the resulting forces will guide the structure back to its equilibrium state during relaxation.
This insight was implicitly used in Crystal Diffusion Variational AutoEncoder (CDVAE)~\cite{xie2021crystal}.
The idea of ``forces'' can be furthur generalized into anyresponse properties of systems to external noises, including the response to lattice deformation.

Here, we introduce a novel generative method for materials and molecules, Response Matching (RM), which leverages the locality of atomic interactions, atomic repulsion, and the PES minimum, while naturally incorporating permutation, translation, rotation, and periodic invariances. We also highlight how RM is closely related to the Denoising Diffusion Probabilistic Model (DDPM)~\cite{ho2020denoising, nichol2021improved}. Finally, we demonstrate the RM model across three systems: a small organic molecular dataset, stable crystals from the Materials Project~\cite{jain2013commentary}, and a single diamond configuration training datum.

\section{Related Work}

\subsection{Generative models of materials and molecules}
Earlier works on generative models of small molecules include the autoregressive G-SchNet~\cite{gebauer2019symmetry} and equivariant normalizing flows (ENF)~\cite{garcia2021n} have been employed to generate three-dimensional equilibrium structures of small organic molecules, while variational autoencoders have been used for material generation~\cite{xie2021crystal}. More recently, diffusion models have been utilized for generating molecules~\cite{hoogeboom2022equivariant,wu2022diffusion,xu2023geometric,peng2023moldiff,huang2023mdm,morehead2023geometry}. Additionally, these models can be conditioned on specific chemical or biological properties through guidance mechanisms~\cite{ragoza2022generating,gebauer2022inverse,corso2022diffdock}.

For generating materials, the PBC 
and the lattice vectors of the cell need to be considered in addition to the atomic coordinates.
Available methods include: latent representations can be directly used ~\cite{pickard2024}, or used together with variational autoencoder to generate stable three-dimensional structures in CDVAE~\cite{xie2021crystal}.
MatterGen performs joint diffusion on atom type, coordinates, and lattice~\cite{zeni2023mattergen}.
A double diffusion model on both lattice and coordinates also pre-selects space groups~\cite{sultanov2023data}.
DiffCSP is a diffusion model for crystal structure prediction purposes that utilizes fractional coordinates~\cite{jiao2024crystal}.

\subsection{Machine learning interatomic potentials}

Machine learning interatomic potentials (MLIPs) enable precise and comprehensive exploration of material and molecular properties at scale, by learning from quantum-mechanical calculations and then predicting the energy and forces of atomic configurations speedily~\cite{keith2021combining,unke2021machine}.
Most MLIPs exploit the nearsightedness of atomic interactions, expressing the total potential energy of the system as the sum of the atomic energies for each atom, i.e.:
\begin{equation}
    E = \sum_i E_i.
    \label{eq:E-mlp}
\end{equation}
The forces can readily be computed by taking the derivatives of the total energy with respect to atomic coordinates.
and the stress can be computed from the virial.

There are many MLIP methods available, e.g., Behler-Parrinello neural network potentals~\cite{behler2007generalized}, GAP~\cite{bartok2010gaussian}, Moment Tensor Potentials (MTPs)~\cite{shapeev2016moment}, Atomic Cluster Expansion (ACE)~\cite{drautz2019atomic}, NequIP~\cite{batzner20223}, MACE~\cite{batatia2022mace}, to name a few.
In this work, we use Cartesian Atomic Cluster Expansion
(CACE)~\cite{cheng2024cartesian} without using message passing layers, due to its efficiency and alchemical learning capabilities, which are important for learning PES of materials with diverse elements.

In CACE, an atom is treated as a node on a graph, and edges connects atoms within a cutoff radius $r_\mathrm{cut}$.
Each chemical element is embedded using a learnable vector $\boldsymbol{\theta}$ with dimension $N_{embedding}$.
The type of edge that connects two atoms, $i$ and $j$, is encoded using the tensor product of the embedding vectors of the two nodes,
$T=\boldsymbol{\theta}_i \otimes \boldsymbol{\theta}_j$.
The length of the edge, $r_{ji}$, is described using a radial basis $R$.
The angular component of the edge, $\hat{\mathbf{r}}_{ji}$, is encoded using an angular basis $L$.
The edge basis combines all this information:
\begin{equation}
    \chi_{cn\mathbf{l}} (i, j) =  
    T_{c}(\boldsymbol{\theta}_i, \boldsymbol{\theta}_j)
    R_{n}(r_{ji})
    L_{\mathbf{l}}(\hat{\mathbf{r}}_{ji}).
    \label{eq:edge-basis}
\end{equation}
The atom-centered representation is made by 
summing over all the edges of a node,
\begin{equation}
    A_{i, cn\mathbf{l}} = 
    \sum_{j\in \mathcal{N}(i)} \chi_{cn\mathbf{l}} (i, j).
\end{equation}
The orientation-dependent $A$ features are symmetrized to get the rotationally-invariant $B$ features of different body orders $\nu$, e.g.
for $\nu = 2$~\cite{Zhang2021}, 
\begin{equation}
    B_{i, cnl}^{(2)} = 
    \sum_{\mathbf{l}} 
    \mathcal{C}(\mathbf{l})
    A_{i, cn\mathbf{l}}^2.
\end{equation}
Then, a multilayer perceptron (MLP) maps these invariant features to the target of the atomic energy of each atom $i$,
\begin{equation}
    E_i = MLP(B_i).
\end{equation}

\subsection{DDPM}
Our method is closely related to the foundational work on DDPMs~\citep{ho2020denoising, nichol2021improved}, which involves a noise model and a denoising neural network. 
The noise model corrupts a data point $\mathbf{x}$ to a sampled log signal-to-noise ratio, $\lambda$, as follows:
\begin{equation}
    \mathbf{x}_{\lambda} = \alpha_{\lambda}\mathbf{x} + \sigma_{\lambda}\mathbf{\epsilon},
\end{equation}
where $\mathbf{\epsilon}$ is a random noise, $\alpha_{\lambda}^2 = 1/(1+e^{-\lambda})$, and $\sigma^2_{\lambda} = 1-\alpha_{\lambda}^2$.
The denoising model learns to predict the clean input $\mathbf{x}$ from $\mathbf{x}_\lambda$, or equivalently, the added noise. 
The denoising neural network with parameters $\theta$ is trained on the score matching objective over multiple noise scales:
\begin{equation}
    L_{\lambda} = \lVert \mathbf{\epsilon}^\theta (\mathbf{x}_\lambda) - \epsilon\rVert^2
    \label{eq:L-DDPM}
\end{equation}

\subsection{Crystal structure prediction}

Crystal structure prediction (CSP) is a computational method that combine quantum mechanical calculations, such as density functional theory (DFT)
with optimization algorithms to search for local minima on the PES. 
Notable examples include USPEX that uses the evolutionary algorithm to search for stable structures~\cite{jiao2024crystal,oganov2006crystal}, and random structure search (RSS) that starts with random atomic positions and then performs PES minimization~\cite{Pickard2011}.
As DFT is computationally expensive, recent studies have started to use MLIPs as the surrogate PES, e.g.~\cite{cheng2022crystal,salzbrenner2023developments,merchant2023scaling}.

\section{Methods}

Just like DDPM, Response Matching also includes a noise step and a denoising step. Noise is directly applied to the Cartesian coordinates of atomic structures. Consider the coordinates of an equilibrium structure as $\textbf{R}_0 = (\textbf{r}_1, ..., \textbf{r}_N)$, where $\textbf{r}_i$ denotes the position of atom $i$. Random atomic displacements are added to this initial structure, generating a sequence of increasingly amorphous structures, $\mathbf{R}_\lambda$. These noisy displacements can be determined all at once based on the chosen value of $\lambda$, rather than being added step by step. The displacement for each atom $i$ is represented as $\Delta \textbf{r}_{i, \lambda}$, and for periodic atomic structures, this displacement follows the minimal image convention in PBC. For simplicity, we later omit $\lambda$ in the subscripts.

As each equilibrium structure lies at the minimum of the PES, the forces on atoms are nearly zero. 
When displacements are applied to these atomic positions, the atomic forces deviate from zero and tend to pull the atoms back to their original coordinates. 
The harmonic approximation, a simple yet widely used physical assumption in quantum mechanical calculations and atomistic modeling, states:
the force on the atom $i$ is
\begin{equation}
    \Tilde{\textbf{F}}_i^{H} = - k \Delta \textbf{r}_i.
\end{equation}
The tilde notation on the force indicates its fictitious nature, as the force constant $k$ is not a physical value but rather a hyperparameter of the RM model. This approximation effectively attaches a harmonic spring between the current position of atom $i$ and its equilibrium position.

We incorporate another physical inductive bias: atoms cannot approach too closely due to strong repulsive forces at short distances. 
To account for this effect, we apply a short-range repulsive pairwise potential between atom pairs $i$ and $j$, such as:
\begin{equation}
g_c(r_{ji}) = 
\begin{cases} 
  m \left(1-\dfrac{r_{ji}^2}{r_c^2}\right)^{n} & \text{if } r_{ji} < r_c \\
  0 & \text{if } r_{ji} \geq r_c,
\end{cases}
\end{equation}
where $r_{ji}$ is the scalar distance between the pair, $r_c$ is typically a fraction of an Angstrom, and $m$ and $n$ are hyperparameters dictating the strength of the repulsion.
The resulting repulsive force on the atom $i$ is 
\begin{equation}
    \Tilde{\textbf{F}}_i^{R} = - \sum_{j\in \mathcal{N}(i)} \dfrac{d g_c(r_{ji})}{d r_i},
\end{equation}
summed over all the atoms $j$ within the distance $r_c$ from the atom $i$.
Combining the harmonic and repulsive forces, the total fictitious force on the atom $i$ is $\Tilde{\textbf{F}}_i = \Tilde{\textbf{F}}_i^{H} + \Tilde{\textbf{F}}_i^{R}$.

For periodic systems, we also add distortions to their periodic cell to create elastic strain $\mathbf{\gamma}$.
The associated stress is approximated utilizing the stress-strain relationship for isotropic elastic materials:
\begin{equation}
    \Tilde{\mathbf{\sigma}} = \mathbf{C} \mathbf{\gamma},
\end{equation}
where the components of \( \mathbf{\epsilon} \) include normal and shear moduli that are treated as hyperparameters.
For molecules without periodicity, the lattice strain step is skipped.

The general idea of RM is to use a denoising model with parameters $\theta$ to fit to the fictitious response properties, i.e., forces and stresses. 
The corresponding objective function is:
\begin{equation}
    L_{\lambda} =  \sum_{i=1}^{N} \|\Tilde{\mathbf{F}}^\theta_i - \Tilde{\mathbf{F}}_i\|^2
    + \beta \|\Tilde{{\sigma}}^\theta- \Tilde{\sigma}\|^2,
    \label{eq:loss}
\end{equation}
where $\beta$ balances the relative weight between the force loss and the stress loss.
Compared with the objective in DDPM (Eqn.~\eqref{eq:L-DDPM}),
One can see that the mathematical framework is identical, with RM employing a physics-inspired response to the noise instead of the noise itself.

Using MLIPs for this denoising model is advantageous because they inherently incorporate the translational, rotational, and permutational symmetries of atomic systems. 
After the MLIP is trained using the objective in Eqn.~\eqref{eq:loss}, it is then used for the denoising step in RM.
As the denoising model has a pseudo potential energy surface $\Tilde{E}$, rather than performing denoising with
a fixed schedule,
one can directly search for local minima on the $\Tilde{E}$.
This is effectively a random structure search process~\cite{Pickard2011}:
For each RSS run, we first chose a reasonable cell shape at random, and added atoms of chosen the elements and composition into the simulation cell at random positions while keeping the initial density of the cell close to the typical density range of this system.
We set a lower bound on the interatomic
distance for each pair of atomic species, but otherwise imposed no additional constraints on the initial structures.
We then relax both the atomic positions and simulation cell using the FIRE (Fast Inertial Relaxation Engine) optimizer~\cite{bitzek2006structural}, which continues until the pseudo forces on the atoms become negligible. FIRE dynamically adjusts step sizes for faster convergence~\cite{bitzek2006structural}, making it a common choice in atomistic simulations due to its efficiency.
Alternatively, one can use other methods such as Langevin dynamics, simulated annealing~\cite{bertsimas1993simulated}, 
evolutionary algorithm~\cite{oganov2006crystal},
particle-swarm optimization and Bayesian optimization~\cite{cheng2022crystal}.
The choice of the optimization method will likely influence the probability of finding the global minimum on the PES versus other local minima.
For the purpose of finding new materials and molecules, it may not always be advantageous to
maximize this probability, as the local minima can also correspond to synthesizable structures.

\section{Examples}

\subsection{Molecule generation: QM7b}

\begin{figure*}
    \centering
    \includegraphics[width=0.99\textwidth]{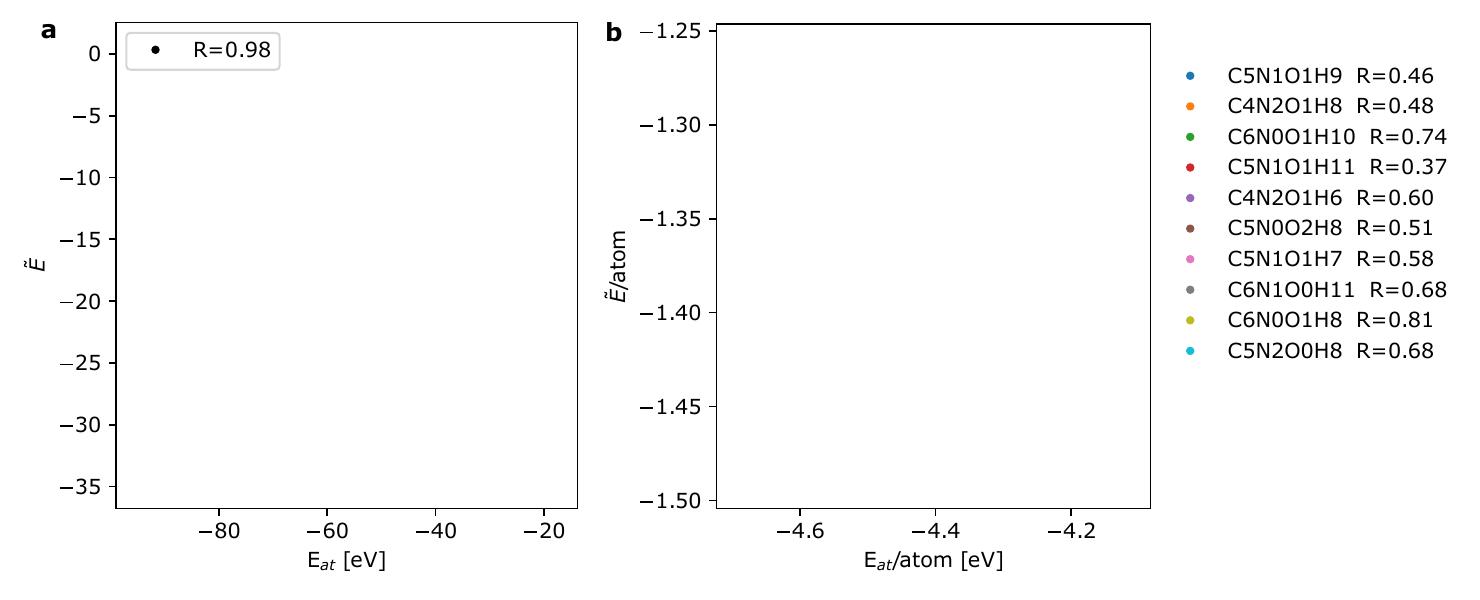}
    \caption{The comparison between the actual atomization energy ($E_{at}$) and the pseudo energy ($\Tilde{E}$) predicted by the RM model for small molecules in the QM7b data set.
    \textbf{a} shows the comparison of the energies per molecule.
    \textbf{b} shows the comparison of the energies per atom for the molecules with most common compositions in QM7b. The Pearson correlation coefficients $R$ are provided in the legends.}
    \label{fig:qm7b-energy}
\end{figure*}

QM7b~\cite{qm7b} is curated based on GDB-13~\cite{gdb} (a database of nearly 1 billion stable and synthetically accessible organic molecules),
and it is composed of 7,211 molecules of up to 23 atoms (including 7 heavy atoms of C, N, O, S, and Cl).

During training, random displacements with a maximum magnitude of 1.6~\AA{} were added to all the atoms in the original molecules, and a CACE potential was employed to learn these pseudo forces. We used $r_\mathrm{cut} = 4.5$~\AA, $l_\mathrm{max} = 3$, $\nu_\mathrm{max} = 3$, and $N_\mathrm{embedding} = 3$. The model has 16,490 trainable parameters.
Training takes roughly one day on a laptop.

Notably, as training is solely based on the pseudo forces, atomization energy or other molecular properties were not included. However, as seen in Fig.~\ref{fig:qm7b-energy}a, $E_{at}$ is highly correlated with $\Tilde{E}$, with a Pearson correlation coefficient of $R = 0.98$. This high correlation primarily results from the additivity of atomic energies (refer to Eqn.~\eqref{eq:E-mlp}), an important bias in the MLIP. Furthermore, the MLIP captures subtle energy differences between distinct atomic environments.
We compared the actual atomization energy per atom with the pseudo energy per atom for molecules with the most common compositions in QM7b. Each composition appears approximately 200-300 times in the training set. Fig.~\ref{fig:qm7b-energy}a illustrates that $E_{at}$ per atom is significantly correlated with $\Tilde{E}$ per atom across these compositions. The MLIP identifies these energy differences because the training structures are PES minima.

\begin{figure}
    \centering
    \includegraphics[width=0.99\textwidth]{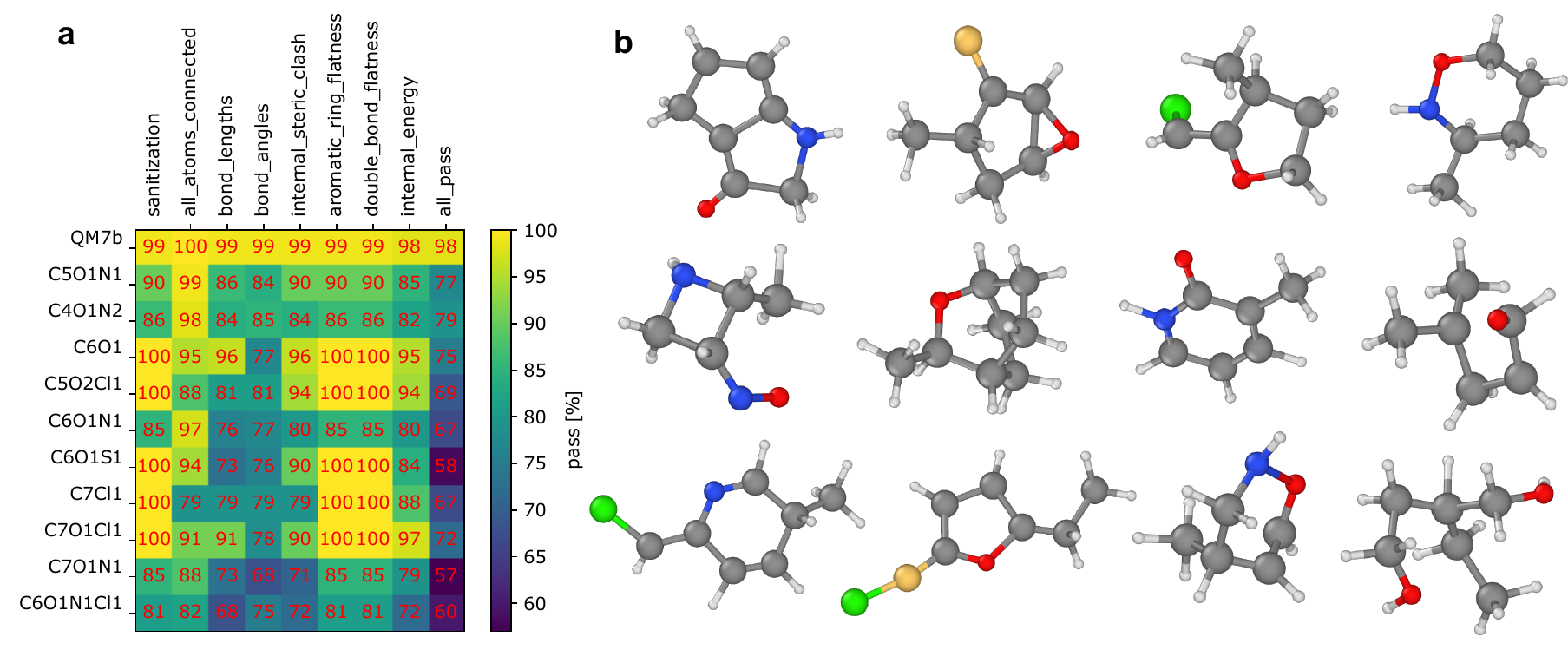}
    \caption{
    Illustrations of small molecules generated using the RM model.
    \textbf{a} shows the percentage of the molecules with given composition that passes the chemical feasibility checks using PoseBusters~\cite{buttenschoen2024posebusters}.
    The last column indicates the percentage that passes all the checks.
    \textbf{b} contains selected molecular configurations.
    The carbon, oxygen, nitrogen, sulfur, chlorine, and hydrogen atoms are colored using black, red, blue, yellow, green, and white, repectively.
    }
    \label{fig:qm7b-checks}
\end{figure}

To generate new molecules, we randomly placed atoms of specific compositions in an orthorhombic box without periodic conditions, with the sole constraint that the minimum interatomic distance be at least 0.7~\AA. We conducted two sets of generation tasks: the first set used compositions well-represented in the training set, while the second set included compositions that were out-of-distribution (containing 8-9 heavy atoms). 
Generating a single structure by geometry relaxation generally takes less than 10 seconds on a laptop.
Currently, the geometry relaxation is performed in serial, and it may be dramatically accelerated by relaxing many molecules together using one, sparsely connected, atomic graph.
Fig.~\ref{fig:qm7b-checks} displays selected configurations generated using the denoising model.

We assess model performance by evaluating the chemical feasibility of the generated molecules, determining whether the model can learn chemical rules from data. Finding a rigorous and unbiased evaluation metric is challenging, and previous studies have used various criteria to assess the feasibility of a molecule represented in three-dimensional coordinates.
For instance, a stability metric~\cite{satorras2021n} checks whether all atoms in a molecule have correct valence based on specific bond distances. A common validity measure~\cite{simonovsky2018graphvae} assesses whether a molecule can be sanitized by RDKit~\cite{bento2020open} using default settings.
According to these metrics, 4.9\% of generated molecules were considered stable and 40.2\% valid using ENF trained on around 14,000 QM9 molecules~\cite{satorras2021n}. Training with 100k QM9 molecules, the E(3) Equivariant Diffusion Model (EDM) achieved 82\% stability and 91.9\% validity, while the MiDi diffusion model, which generates both 2D molecular graphs and their corresponding 3D coordinates~\cite{vignac2023midi}, reached 84\% stability and 97.9\% validity.
Furthermore, the Geometric Latent Diffusion Model (GEOLDM)~\cite{xu2023geometric} attained 89.4\% stability and 93.8\% validity, and the Geometry-Complete Diffusion Model (GCDM)~\cite{morehead2023geometry} achieved 85.7\% stability and 94.8\% validity.

Here we use a set of comprehensive and stringent criteria offered by
PoseBusters~\cite{buttenschoen2024posebusters}: 
RDKit’s chemical sanitisation check (equivalent to the aforementioned validity),  
all atoms connected,
bond lengths,
bond angles,
internal steric clash,   
aromatic ring flatness,
double bond flatness, and the calculated energy of the input molecule based on a forcefield.
The energy calculation step also includes a valency check for each atom, akin to the stability check.

Fig.~\ref{fig:qm7b-checks} illustrates the percentage of molecules with given compositions that pass these chemical feasibility checks. For comparison, the first row displays statistics for the QM7b set. Each criterion provides different insights into the chemical structure, and the pass percentages vary but are correlated. The feasibility percentages are highly dependent on specific compositions.
Overall, the feasibility rate is quite high. The sanitization (validity) rate ranges from 80\% to 100\%, which is similar to rates reported in previous studies trained on the QM9 dataset.

\subsection{Material generation: Materials Project structures}

We used the MP-20 structures as in Ref.~\cite{xie2021crystal}, includes almost all experimentally stable materials from the Materials Project (MP)~\cite{jain2013commentary} with unit cells including at most 20 atoms. The training set contains 27,136 structures.

During the training, cell distortion of up to 0.1 was applied, and random displacements with a maximum magnitude of 0.8~\AA{} were added to all atoms. 
A CACE potential was used to learn the these pseudo forces and stresses.
We used $r_\mathrm{cut}=4$~\AA, $l_\mathrm{max}=2$, $\nu_\mathrm{max}=2$, and $N_\mathrm{embedding} = 4$ for aggressive alchemical compression.
The model has 16,312 trainable parameters.
The training time took two days on an A10 card.

\begin{figure*}
    \centering
    \includegraphics[width=0.95\textwidth]{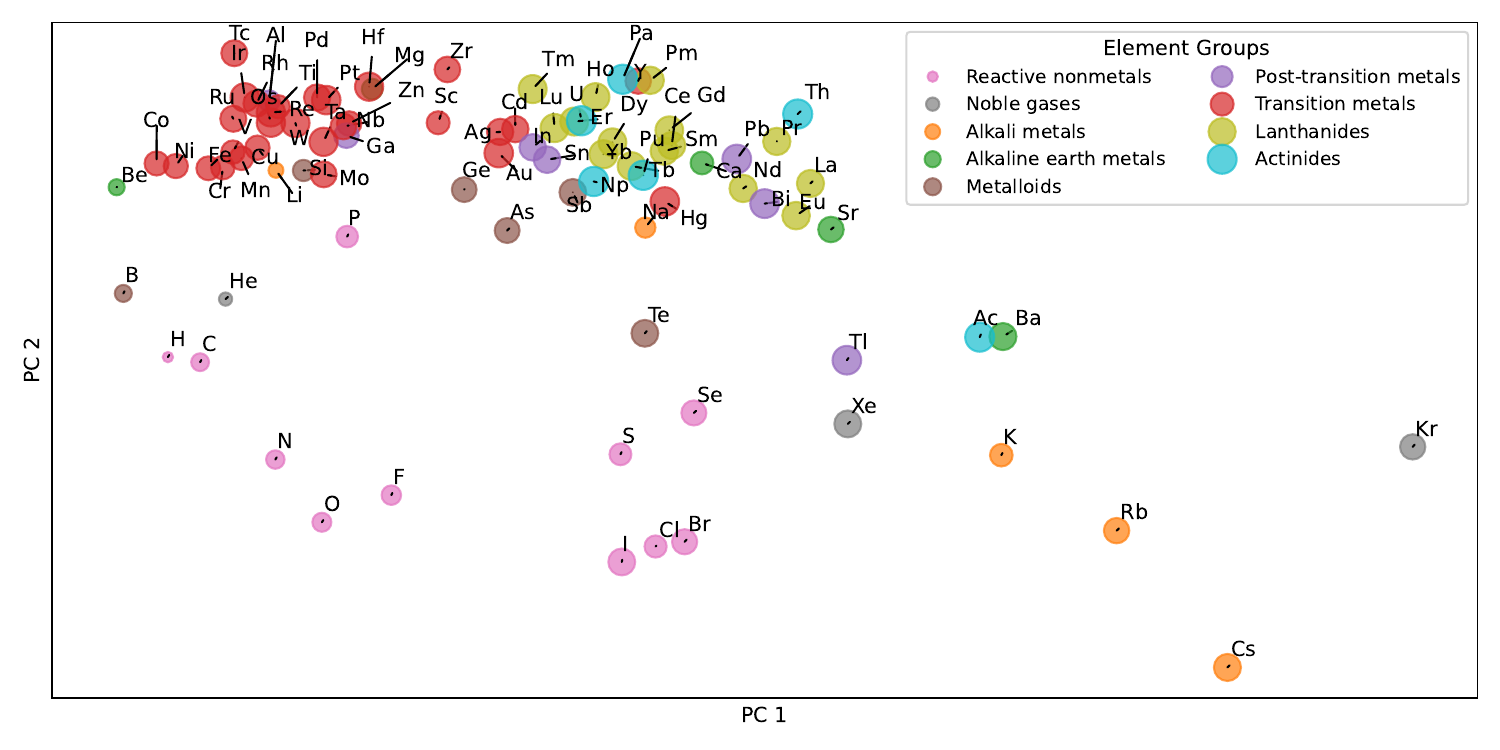}
    \caption{
    The similarity of chemical elements visualized using the first two principal components (PCs) of the CACE embedding matrix $\theta$ in the RM denoising model. Each element is colored according to its chemical group. The size of the symbol indicate the size of the elements. The noble gas element Ne is outside the plot.
    }
    \label{fig:mp-embedding}
\end{figure*}

The alchemical learning capacity of CACE not only enhances learning efficiency, but is crucial to developing a model that is applicable throughout the periodic table. The learnable embedding $\boldsymbol{\theta}$ for each element type encodes its chemical information and can be visualized to provide insights into data-driven similarities.
Given that $N_\mathrm{embedding}=4$ in this case, we performed a principal component analysis (PCA) and plotted the first two principal component axes in Fig.~\ref{fig:mp-embedding}. The elements are color-coded on the basis of their chemical groups, and the PCA map reveals that elements in the same group tend to cluster together. Elements within a group often have similar appearances and behaviors because they possess the same number of electrons in their outermost shell. This demonstrates that the element embedding scheme effectively captures the nature of the periodic table in a data-driven manner.

To generate new crystals,
we randomly placed atoms of specific compositions in a box with periodic conditions, 
with an initial molar volume that is from linear regression of the total volume with respect to the chemical elements from training structures.
We then performed FIRE optimization both for the cell and atomic positions.
We test our model on the same selected set as used in DiffCSP~\cite{jiao2024crystal}, which contains 10 binary and 5 ternary compounds in MP-20 test set. 
For each of these compositions, we generate 30 structures and compared them to the group truth using StructureMatcher from pymatgen~\cite{ong2013python}. 
For eight of these compounds (Ag$_6$O$_2$, Bi$_2$F$_8$, Co$_2$Sb$_2$, Co$_4$B$_2$, Cr$_4$Si$_4$, KZnF$_3$, Sr$_2$O$_4$, YMg$_3$), we found the ground truth structures.
Such match rate is similar to USPEX~\cite{jiao2024crystal,oganov2006crystal}, a CSP method based on DFT,
although worse than the 11/15 match rate using DiffCSP~\cite{jiao2024crystal}.

\begin{figure}
    \centering
   \includegraphics[width=0.6\textwidth]{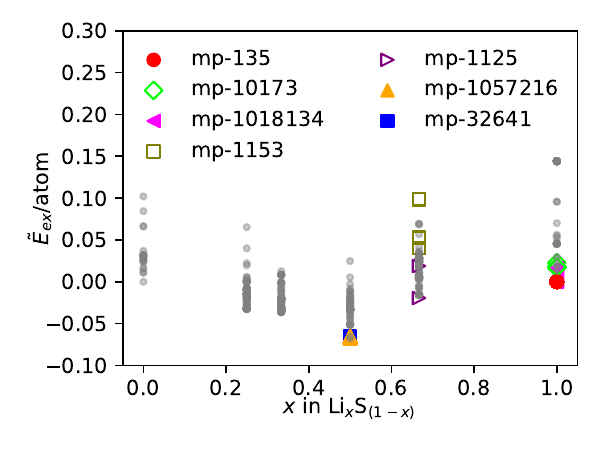}
    \caption{The pseudo convex hull of the generated Li-S structures at different Li fractions. The colored markers denote the generated structures that matched with the known structures in the Materials Project~\cite{jain2013commentary} using StructureMatcher from pymatgen~\cite{ong2013python}, and the materials IDs are given in the legend.
    The filled (hollow) symbols indicate structures that are in (not in) the training set.}
    \label{fig:LiS}
\end{figure}

To demonstrate how the FM model can help scientific applications,
we used it for the
Li-S battery system, which is an attractive candidate for numerous energy storage applications~\cite{see2014ab}.
We searched for a number of stoichiometries Li$_x$S$_{(1-x)}$. As sulfur is highly polymorphic and has complex structures, we used the known pure S structures from MP. 
To organize the generated structures, we plotted the pseudo convex hull using the pseudo excess energy, $\Tilde{E}_{ex} = \Tilde{E} - x\Tilde{E}_\mathrm{Li}-(1-x)\Tilde{E}_\mathrm{S}$, in Fig.~\ref{fig:LiS}.
In reality, Li$_2$S should be the stable phase on the energy convex hall~\cite{see2014ab} rather than LiS, but the overall trend in stability for LiS compounds is captured.
It is worth cautioning that such results are not trustworthy, especially because energies did not enter the training of the model, and one always has to check against quantum mechanical calculations.
Nevertheless, the trained RM model can offer a quick and perhaps insightful first step in exploring a system.
Several structures in the MP~\cite{jain2013commentary} are found,
and they typically have lower pseudo energies, which suggests that one can use $\Tilde{E}$ for additional screening.
Among the seven matched MP structures, only the materials with IDs mp-10173 (Li, P6$_3$/mmc), mp-1125 (Li$_2$S, Pnma), mp-1153 (Li$_2$S, Fm3$\bar{m}$) are in the training set.

\subsection{One-shot learning: A single diamond structure}

To show data efficiency and generalization of the RM,
we trained on a single data point of a cubic diamond structure.
In the noising stage,
uniform random displacement with a maximum magnitude of 0.8~\AA, and lattice strain up to 0.1 were added to the original structure with a molar volume of 4.4~\AA$^3$.
For the CACE potential, we used $r_\mathrm{cut}=4.5$~\AA, $l_\mathrm{max}=3$, $\nu_\mathrm{max}=3$, and $N_\mathrm{embedding} = 1$.
The model only has 2,137 trainable parameters.
Training takes less than an hour on a laptop.
During the relaxation stage, 2-12 carbon atoms were randomly placed in a simulation box with the molar volume set to be between 3.8~\AA$^3$ and 5.6~\AA$^3$.
Generating one structure takes a few seconds depending on the system size.

\begin{figure}
    \centering
   \includegraphics[width=0.6\textwidth]{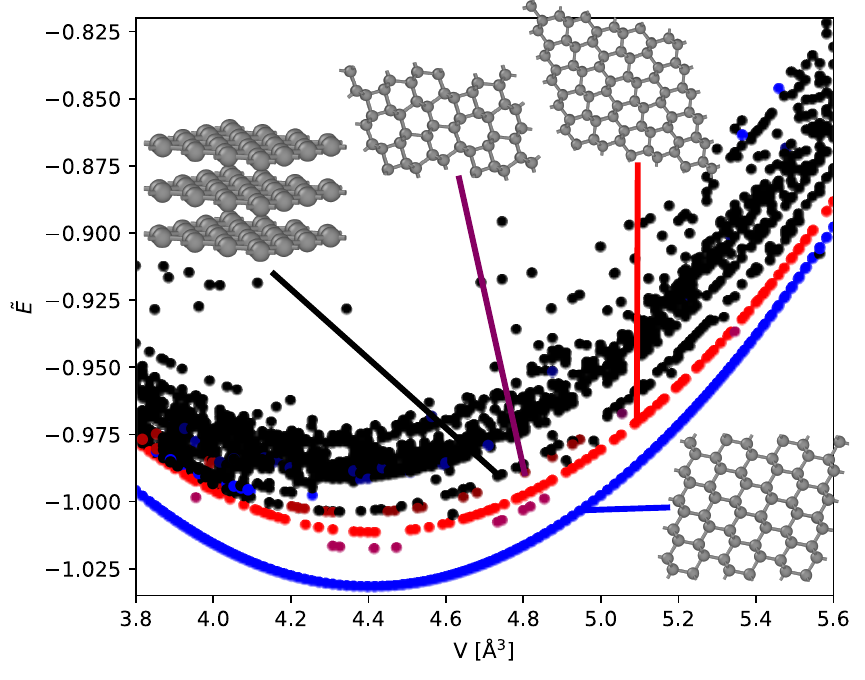}
    \caption{The pseudo energy of the generated carbon structures with different molar volumes. The blue markers represent cubic diamond structures, the red markers indicate hexagonal diamonds, and the purple markers denote diamonds with stacking faults. Graphite structures lie along the band pointed to by the black line.}
    \label{fig:diamond}
\end{figure}

We plot the pseudo-energy $\Tilde{E}$ against molar volume for the obtained structures in Fig.~\ref{fig:diamond}. The cubic diamond structures (blue dots in Fig.~\ref{fig:diamond}) consistently have the lowest $\Tilde{E}$ at all molar volumes compared to the other structures found. The lowest-energy cubic diamond has a molar volume of 4.4 \AA$^3$, which matches the training configuration exactly.
Hexagonal diamonds (red dots in Fig.~\ref{fig:diamond}) and diamonds with stacking faults (purple dots in Fig.~\ref{fig:diamond}) are also frequently found. Notably, a set of graphite structures (in the inset of Fig.~\ref{fig:diamond}) with varying volumes were identified. Graphite structures differ significantly from diamonds, consisting of stacked layers of carbon atoms in a hexagonal lattice. Within each layer, carbon atoms form strong covalent bonds with three neighboring atoms in a trigonal planar arrangement.
The hexagonal diamonds, diamonds with stacking faults, and graphite structures are all local minima on the pseudo energy surface, while the cubic diamond is the true global minimu.
This example thus demonstrates that tracing the local minima can generate out-of-distribution yet physically meaningful structures.

\section{Discussion}

In summary, Response Matching (RM) is a novel generative method that has a noising step equivalent to other diffusion models, and a denoising model that is effectively crystal structure prediction using a machine learning interatomic potential. Just as DDPMs~\citep{ho2020denoising, nichol2021improved} are trained to predict added noise, the MLIP in RM is trained to predict the response of the system to the added noise. 
These responses, in the form of pseudo forces and stress, are simply proportional to the noise.
The benefits of using MLIP for denoising include:
(i) Exploiting the locality of atomic interactions while naturally respecting permutation, translation, rotation, and periodic invariances.
(ii) Allowing RM to simultaneously handle both molecules and bulk materials with and without periodic boundary conditions.
(iii) Enabling advanced optimization methods such as FIRE~\cite{bitzek2006structural}, simulated annealing~\cite{bertsimas1993simulated}, particle-swarm optimization and Bayesian optimization~\cite{cheng2022crystal} during denoising rather than adhering to a fixed schedule.
(iv) The pesudo energies from the RM model are not directly trained but are somehow correlated with the real energies of the systems,
which may offer physical insights and additional way of screening the generated structures.
Moreover, given that MLIPs are well-developed, their advances are directly transferable to generative models of materials and molecules.


\section{Limitations}

The current RM model can be extended in several ways. 
(i) Most crystals fall into a limited set of space groups. In crystal structure prediction, a common technique is to select a space group based on popularity and snap to it during relaxation~\cite{Pickard2011}. 
This approach can also be applied to RM, which could enhance model efficiency by reducing the search space to realistic structures and facilitating the generation of experimentally relevant crystals. 
(ii) One can incorporate another term, $\|y^\theta- y\|^2$ in the objective function in Eqn.~\eqref{eq:loss}, with $y$ being the specific properties, such as bandgap, solubility, or thermal stability. 
This can enable RM to generate molecules and materials conditioned on specific properties, which is crucial for designing functional materials with targeted properties. 
(iii) Rather than fixing composition, alchemical swaps of elements using Monte Carlo moves or another generative model can dynamically determine atom types during denoising. 
(iv) One can further synergize RM and MLIPs, e.g. 
training foundation models using both data from quantum mechanical calculations and unlabelled structural data.
These enhancements would further expand the applicability and flexibility of RM in the generation of materials and molecular structures.

\end{document}